\documentclass[conference]{IEEEtran}
\IEEEoverridecommandlockouts
\usepackage{amsmath,amssymb,amsfonts}
\usepackage{amstext}
\usepackage{breqn}
\usepackage{algorithmic}
\usepackage{graphicx}
\usepackage{textcomp}
\usepackage{xcolor}
\usepackage{hyperref}
\usepackage{todonotes}
\usepackage{algorithm}
\usepackage{algorithmic}
\usepackage{bm}
\usepackage{amsthm}
\DeclareMathOperator*{\minimize}{minimize}

\def\BibTeX{{\rm B\kern-.05em{\sc i\kern-.025em b}\kern-.08em
    T\kern-.1667em\lower.7ex\hbox{E}\kern-.125emX}}
\begin{document}

\title{Two-Memory Reinforcement Learning}

\author{\IEEEauthorblockN{Zhao Yang, Thomas. M. Moerland, Mike Preuss, Aske Plaat}
%\author{\IEEEauthorblockN{Anonymous submission}
\IEEEauthorblockA{\textit{Leiden Institute of Advanced Computer Science } \\
\textit{Leiden University}\\
z.yang@liacs.leidenuniv.nl}
}

\maketitle

\begin{abstract}
While deep reinforcement learning has shown important empirical success, it tends to learn relatively slow due to slow propagation of rewards information and slow update of parametric neural networks. Non-parametric episodic memory, on the other hand, provides a faster learning alternative that does not require representation learning and uses maximum episodic return as state-action values for action selection. Episodic memory and reinforcement learning both have their own strengths and weaknesses. Notably, humans can leverage multiple memory systems concurrently during learning and benefit from all of them. In this work, we propose a method called Two-Memory reinforcement learning agent ({\it 2M})\footnote{Code will be available after authors notification.} that combines episodic memory and reinforcement learning to distill both of their strengths. The {\it 2M} agent exploits the speed of the episodic memory part and the optimality and the generalization capacity of the reinforcement learning part to complement each other. Our experiments demonstrate that the {\it 2M} agent is more data efficient and outperforms both pure episodic memory and pure reinforcement learning, as well as a state-of-the-art memory-augmented RL agent. Moreover, the proposed approach provides a general framework that can be used to combine any episodic memory agent with other off-policy reinforcement learning algorithms.
%\footnote{Please see the anonymous code \href{https://www.dropbox.com/sh/zds4qba8k57isx7/AAB24jcV3dZsNdS_2JYPHa-va?dl=0}{here}.}
\end{abstract}

%%%%%%%%%%%%%%%%%%%%%%%%%%%%%%%%%%%%%%%%%%%%
\section{Introduction}
Deep reinforcement learning (DRL) achieves impressive results in a wide range of domains. It reaches super-human performance in games such as Atari~\cite{mnih2013playing}, Go~\cite{silver2017mastering} and Gran Turismo~\cite{wurman2022outracing}. Recently, it also has shown promise in scientific applications such as controlling  nuclear plasma fusion~\cite{degrave2022magnetic} and discovering new matrix multiplication algorithms~\cite{fawzi2022discovering}. However, DRL is well-known for being data inefficient, since back-propagation of reward signals and learning updates (including representation learning) can be slow.

In contrast to such parametric learning approaches, non-parametric episodic memory approaches maintain a memory buffer to store high-rewarded trajectories for either action selection~\cite{blundell2016model,pritzel2017neural} or for enhancing other reinforcement learning methods~\cite{lin2018episodic,kuznetsov2021solving,hu2021generalizable}. These papers have been demonstrated to outperform conventional reinforcement learning methods in certain tasks, such as Atari~\cite{bellemare2013arcade}, and Labyrinth~\cite{mnih2016asynchronous}. In episodic memory, reward signals are back-propagated considerably faster than in  one-step temporal difference (TD) learning that is commonly used in reinforcement learning. In addition, one-step TD learning is further slowed down by representation learning and the use of function approximation. A potential problem of episodic memory is that fast back-propagation is problematic in stochastic tasks, and the lack of learnable feature representations can make generalization difficult in episodic memory. The question therefore becomes: can we combine the best of both approaches in a single algorithm?  

Evidence from neuroscience shows that multiple memory systems are activated when humans are learning, and these also interact with each other~\cite{poldrack2003competition,schott2005redefining}.
\begin{figure}
    \centering
    \includegraphics[scale=0.25]{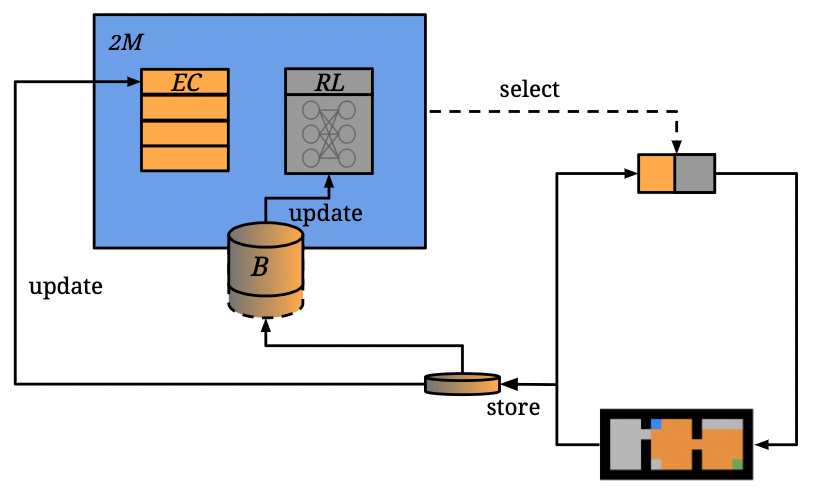}
    \caption{The workflow of the {\it 2M} agent. Before the episode starts, the {\it 2M} agent selects one type of memory for action selection in the next episode, which could either be episodic control (EC) or parametric reinforcement learning (RL). The data subsequently collected is used to update the EC solution (directly), and also enters an experience replay buffer ($B$) for future updating of the parametric RL solution. }
    \label{fig:2Mworkflow}
\end{figure} 
Previous research~\cite{lin2018episodic,kuznetsov2021solving,hu2021generalizable} has shown that integrating episodic memory and reinforcement learning can improve overall performance. In these works, episodic memory is mainly used to provide learning signals for DRL methods, but they again face the same challenges that are inherent to DRL. To fully capitalize on the advantages of episodic memory and reinforcement learning, we propose a novel approach called Two-Memory reinforcement learning ({\it 2M}), in which both approaches complement eachother.

The workflow of the {\it 2M} agent is shown in Fig.~\ref{fig:2Mworkflow}. The {\it 2M} agent maintains two memories, namely `episodic control' (episodic memory for control) ({\it 2M-EC}) and `reinforcement learning' ({\it 2M-RL}). In the beginning, the {\it 2M} agent decides which memory to employ for action selection in the upcoming episode. Then the collected episodic data is pushed into the experience replay buffer where data for training {\it 2M-RL} is sampled from occasionally. Meanwhile, episodic trajectories are used to update {\it 2M-EC}.

The intuition is that after {\it 2M-EC} discovers high-reward trajectories, the {\it 2M} agent is able to retrieve these trajectories quickly although they might be suboptimal. However, at this stage, {\it 2M-RL} still has not learned anything substantial  due to the one-step reward backup and the slow updates required with function approximation (in DRL). Thus, the {\it 2M} agent should prefer to use {\it 2M-EC} initially. As the amount of data collected increases and  training continues, {\it 2M-RL} becomes increasingly good at dealing with stochasticity and developing better feature representations; the {\it 2M} agent should then gradually switch to  {\it 2M-RL}. This is the conceptual approach we study in this work. 

In short, our work makes two main contributions:
\begin{itemize}
    \item We propose a novel framework called {\it 2M} that combines episodic control and reinforcement learning methods, exploiting the strengths from both sides. The framework can be used with any type of EC method and any type of off-policy RL method. 
    \item We conduct experiments showing that {\it 2M} outperforms state-of-the-art baselines, and we also include ablation studies that examine when the {\it 2M} works well, and where it may still be improved.
\end{itemize}

\section{Background}
We will first introduce the formal problem definition and the two main approaches combined in this paper: parametric reinforcement learning (in the form of deep Q-learning) and non-parametric episodic memory. 

\label{se}
\subsection{Markov Decision Process} 
We follow standard definitions~\cite{sutton2018reinforcement} and define decision making problems as a Markov Decision Process (MDP) represented by $\langle S, A, R, P, \gamma \rangle$. $S$ denotes the state space, $A$ denotes the action space, $R$ denotes the reward function and $P$ denotes the dynamic transition. $\gamma$ is the discount factor, which is usually close to 1. The agent interacts with the environment by taking action $a_t\in A$ according to some policy $\pi$ given the current state $s_t\in S$ at time step $t$, then the next state $s_{t+1}\sim P(\cdot|s_t,a_t)$ and reward $r_t=R(s_t, a_t, s_{t+1})$ is returned by the environment. A policy $\pi$ maps a state to a distribution over actions. The agent will then take the next action $a_{t+1}$ based on the new state $s_{t+1}$. This process repeats until the terminal time step $T$. The goal of the agent is to learn a policy $\pi$ that maximizes the expected cumulative reward: $\mathbb{E}_{s_{t+1}\sim P(\cdot|s_t,a_t), a_t\sim \pi(\cdot|s_t), r_t\sim R(\cdot|s_t,a_t)}[\sum_{t=0}^{T}\gamma^{t} \cdot r_t]$.
\subsection{Deep Q-Learning} 
Define the state-action value function $Q(s_t,a_t)$ as the expected cumulative reward the agent will receive by starting from state $s_t$ and taking action $a_t$. Q-learning is an algorithm to learn the optimal Q value function, it updates the state-action value function by iteratively following the Bellman equation. The update rule of Q-learning is:
\begin{equation*}
    Q(s_t, a_t) \gets Q(s_t, a_t) + \alpha (r_t + \gamma \max_{a'\in A}Q(s_{t+1}, a')-Q(s_t, a_t))
\end{equation*}where $\alpha$ is the learning rate and $\gamma$ is the discount factor.
The solution of Q-learning is generally stored in a table. When the state space is large, storing all possible states becomes infeasible. We may then use a neural network to approximate the state-action value function. The update rule of deep Q network~\cite{silver2017mastering} is: 
\begin{equation*}
    y(s_t) \gets r_{t-1} + \gamma \max_{a'\in A}Q(s_{t}, a')
\end{equation*}
\begin{equation}
    \minimize{\mathbb{E}_{s_t, a_t, r_t, s_{t+1} \in D} (y(s_{t+1}) - Q(s_t, a_t))^2}
    \label{eq:dqn}
\end{equation}
where $D$ is the sampled data for training. After we have the state-action value function, the policy is exacted by always taking the action with the highest state-action value greedily in every state.
\subsection{Episodic Control} 
Episodic control refers to a class of methods that directly use non-parametric episodic memory for action selection~\cite{blundell2016model,pritzel2017neural}. It is generally implemented as a table ($Q^{ec}$), and rows represent different actions while columns represent different states. Each entry is associated with a state-action ($s_t, a_t$) pair and denotes the highest encountered episodic return ($G_t=\sum_{k=t}^{T}\gamma^{k-t}r_k$) after taking action $a_t$ in the state $s_t$. The update only occurs after one episode has terminated, which is similar to tabular Monte-Carlo back-up but with a more aggressive update rule. Instead of incrementally updating state-action values, episodic control replaces the stored episodic return with the best value observed so far. The update rule of episodic control is:
\begin{flalign}
 & Q^{ec}(s_t, a_t) \gets
    \begin{cases}
      G_t & \text{if $(s_t, a_t) \notin Q_{ec}$}, \\
      \max \{G_t, Q^{ec}(s_t, a_t)\} & \text{otherwise}
    \end{cases} \nonumber \\
    \label{eq:ec_update}
\end{flalign}
During action selection, if the queried state already has $Q$ values for all actions, we take the action that is with the highest $Q$ value. Otherwise, missed $Q$ values are estimated by averaging over its K-nearest neighbors' $Q$ values. When the memory is full, the least updated entry will be dropped from the table. 
%%%%%%%%%%%%%%%%%%%%%%%%%%%%%%%%%%%%%%%%%%%%
\section{Related Work}
We will first discuss previous work on episodic control, and how it has been used as a training target for deep reinforcement learning. This last approach differs from our work, where EC is used for action selection (see Fig.~\ref{fig:2Mworkflow}). Afterwards, we also briefly discuss related work on experience replay, since it plays a central role in our approach as well (see Fig.~\ref{fig:2Mworkflow}). 

\subsection{Episodic Control}
Model Free Episodic Control (MFEC)~\cite{blundell2016model} is the first episodic control method and it is implemented as a table with untrainable features. Thus, it enjoys limited generalization over either a randomly projected or pre-trained feature space. Neural Episodic Control (NEC)~\cite{pritzel2017neural} solves this limitation by maintaining a differentiable table to learn features while using the same update and action selection rule (shown in Eq.~\ref{eq:ec_update}) as MFEC but achieves better performance. Since Monte-Carlo returns must be stored for each state-action pair, episodic control methods can not deal with continuous action spaces naturally. Continuous Episodic Control (CEC)~\cite{yang2022continuous} extends episodic memory to select actions directly in continuous action space. The {\it 2M} agent integrates MFEC as a core component, and partially uses it for action selection.

\subsection{Episodic Memory for Learning} While using episodic memory for control is fast, reinforcement learning methods might still be preferred in the long run due to their strength. Many approaches use the returns stored in episodic memory to provide richer learning targets for reinforcement learning methods. Episodic Memory Deep Q Network (EMDQN)~\cite{lin2018episodic} uses returns in episodic memory to enhance learning targets in deep Q-Learning. Episodic Memory Actor-Critic (EMAC)~\cite{kuznetsov2021solving} and Generalizable Episodic Memory (GEM)~\cite{hu2021generalizable} uses episodic returns to enhance learning targets in actor-critic methods to solve tasks with continuous action space. Episodic Backward Update (EBU)~\cite{lee2019sample} utilizes structural information of states that are in the same episode and executes a one-step backup for each state along the trajectory. The aforementioned methods take advantage from richer learning signals of episodic memory, but underlying neural network training still progresses slowly. Thereby, the benefit of the EC solution will not affect action selection quickly. In contrast, the {\it 2M} agent does use episodic memory for action selection, which may give it a fast head-start. As a second difference, in {\it 2M} the collected data is also used to train the {\it 2M-RL} agent (in contrast to previous methods). 

\subsection{Experience Replay}
Experience replay was originally proposed to improve data efficiency and break correlations of training data for off-policy reinforcement learning methods. Uniform sampling is the most naive and commonly used way to sample data from the replay buffer for training, where transitions are sampled at the same frequency as they were experienced regardless of their significance~\cite{schaul2015prioritized}. To address this limitation, prioritized Experience Replay (PER)~\cite{schaul2015prioritized} prioritizes transitions that have larger TD errors during the training, and samples these transitions more often because larger TD errors indicate there is more information to learn. Hindsight Experience Replay (HER)~\cite{andrychowicz2017hindsight} is proposed for the multi-goal/goal-conditioned RL setting; it treats states that the agent actually achieves as desired goals and learns from failures. Since not all failures are equally important, Curriculum-guided Hindsight Experience Replay (CHER)~\cite{fang2019curriculum} adaptively replays failed experiences according to their similarities to true goals. An event table~\cite{kompella2022event} is defined as a buffer to store transitions related to important events. The authors theoretically proved that sampling more data from the event table will lead to better performance. Although the {\it 2M} agent doesn't employ any special sampling strategies to sample from the replay buffer, the data stored in the buffer is actually from two different sources ({\it 2M-EC} and {\it 2M-RL}). We can thus vary sampling preference by pushing different amounts of data from different sources.

%%%%%%%%%%%%%%%%%%%%%%%%%%%%%%%%%%%%%%%%%%%%
\section{Two-Memory Reinforcement Learning ({\it 2M})}
We will now formally introduce the {\it 2M} framework. It consists of two `memories', where one represents a fast learning memory (episodic control agent) and another one represents a slow learning memory (reinforcement learning agent, in our work we use 1-step (deep) Q-learning). Intuitively, we should prefer to use the fast (but sub-optimal) learning memory initially, then gradually switch to the slow (but with better asymptotic performance) learning memory. In this section, we will first motivate this intuition by using a very simple example (Sec. \ref{subsec:motiv}). Then we explain the designs of the proposed approach. Since we combine two different methods and want to switch between them, we also need to discuss a scheduling mechanism to decide when to switch (section~\ref{subsec:switch}) and need to decide how to utilize collected data (section~\ref{subsec:learning}). The overall algorithm is detailed in Alg.~\ref{alg:2m}.
\begin{figure}[!htb]
    \centering
    \includegraphics[scale=0.27]{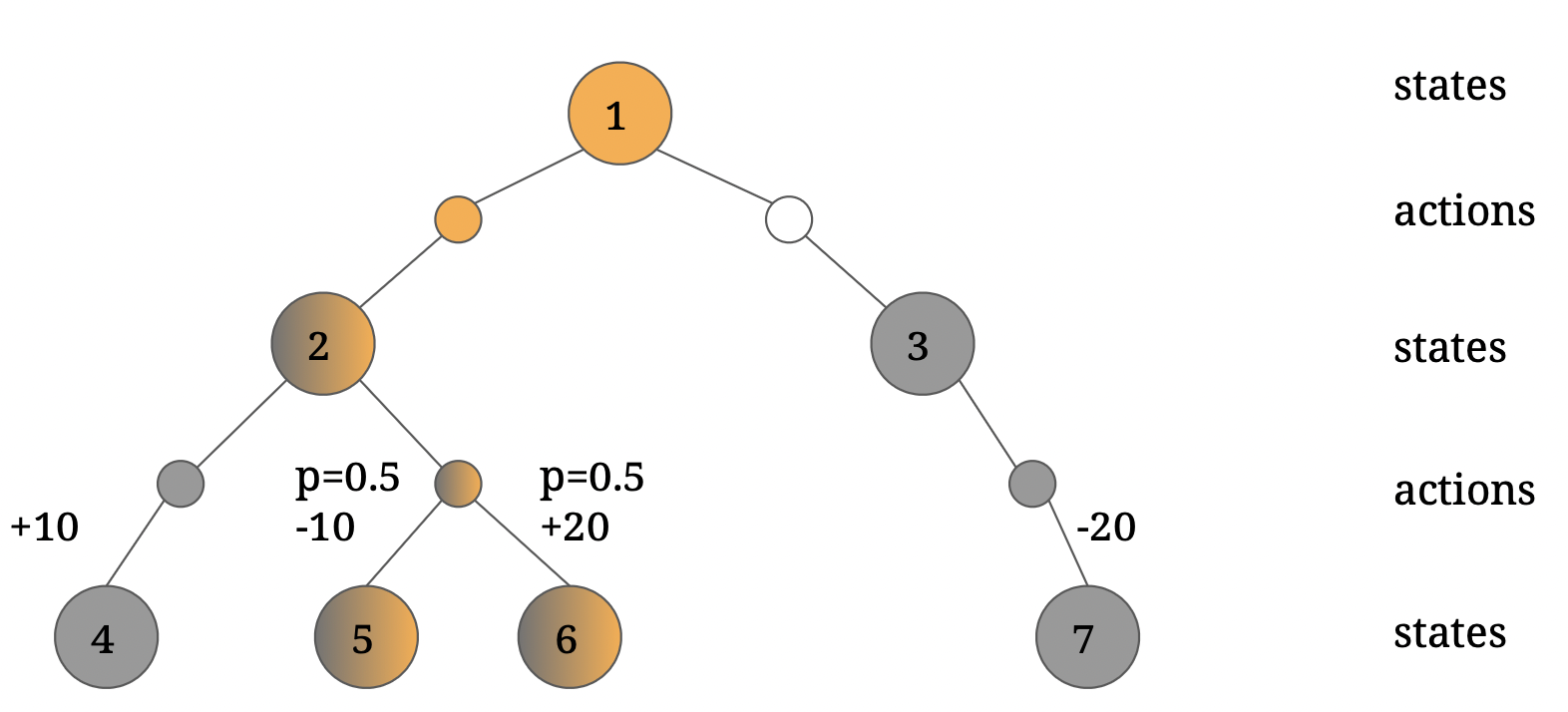}
    \caption{A motivating example with seven states in the state space and two actions in the action space. After the agent discovers all possible trajectories (from $s^1$ to all possible terminal states), EC (orange) finds the sub-optimal solution while one-step off-policy RL (grey) only back-propagates reward signals to direct predecessor states and thus does not change the estimates at the root state. However, after many iterations, off-policy RL will converge to the true optimal values (preferring action left followed by action left), while EC will commit to a suboptimal solution (action left followed by action right). Colour shading indicates the preferred solution path for each method after a single visit to every possible path.}
    \label{fig:motivation}
\end{figure}
\begin{algorithm}[!ht]
   \caption{Two-Memory Reinforcement Learning}
   \label{alg:2m}
\begin{algorithmic}
   %\STATE {\bfseries Input:} data $x_i$, size $m$
   \STATE {\bfseries Input:} Environment \textit{env}, non-parametric episodic control agent \textit{ec}, parameterized reinforcement learning agent \textit{rl}, Two-memory agent {\it 2M}, replay buffer \textit{$B$}, probability $p^{ec}$, exploration factor $\epsilon$
   \STATE $G^{ec} \gets [\ ]$ \hfill \COMMENT{\texttt{initialize return list of ec}}
   \STATE $G^{rl} \gets [\ ]$ \hfill \COMMENT{\texttt{initialize return list of rl}}
   \STATE {\it 2M-RL} $\gets$ {\it rl}
   \STATE {\it 2M-EC} $\gets$ {\it ec}
   \WHILE{training budget left}
   \STATE $s \gets $ reset the \textit{env}
   \STATE {\it done} = false
   \STATE $\tau \gets [\ ]$ \hfill \COMMENT{\texttt{initialize episodic trajectory}}
   \IF{$\Delta\sim U(0,1)<p^{ec}$} 
   \STATE {\it 2M} $\gets$ {\it 2M-EC} \hfill \COMMENT{\texttt{2M switches to EC}}
   \ELSE{\STATE {\it 2M} $\gets$ {\it 2M-RL} \hfill \COMMENT{\texttt{2M switches to RL}}}
   \ENDIF
   \STATE $G \gets 0$
   \WHILE{\textbf{not} {\it done}}
   \STATE execute action $a$ selected by \textit{2M} based on $s$ with $\epsilon$-greedy exploration
   \STATE observe reward $r$ and next state $s'$ in $env$
   \STATE store $(s, a, r, s')$ to $B$ and $\tau$
   \STATE $s \gets s'$
   \STATE $G \gets G + r$
   \STATE update {\it 2M-RL} using $D$ according to (\ref{eq:dqn}) if needed, $D \sim B$
   \ENDWHILE
   \STATE update {\it 2M-RL} using $\tau$ according to (\ref{eq:ec_update})
   \IF{{\it 2M} is {\it 2M-RL}}
   \STATE add $G$ to $G^{rl}$
   \ELSE{\STATE add $G$ to $G^{ec}$}
   \ENDIF
   \STATE update $p^{ec}$ \hfill \COMMENT{\texttt{decay or increase or consistent}}
\ENDWHILE
\end{algorithmic}
\end{algorithm}

\begin{figure*}
    \centering
    \includegraphics[scale=0.5]{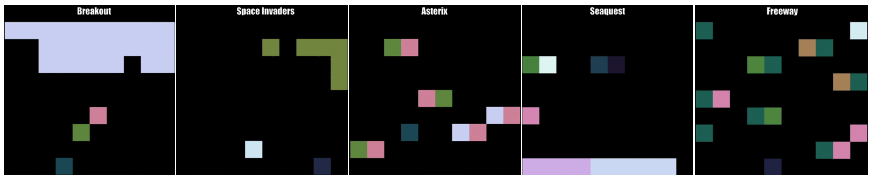}
    \caption{The five MinAtar games used in the experiments (from left to right): Breakout, Space Invaders, Asterix, Seaquest, Freeway.}
    \label{fig:minatar_env}
\end{figure*}

\subsection{A Motivating Example}
\label{subsec:motiv}
We first consider a very simple example domain shown in Fig.~\ref{fig:motivation}, circles with numbers represent different states while smaller circles represent actions. There are seven states in the state space: $S=\{s^1,s^2,s^3,s^4,s^5,s^6,s^7\}$, two actions (left and right) in the action space: $A=\{a^1, a^2\}$. Most of dynamic transitions are deterministic, except in state $s^2$, after taking action $a^2$, there is $50\%$ probability the next state ends up with $s^5$ and $50\%$ probability ends up with $s^6$: $P(s^2,a^2,s^5)=P(s^2,a^2,s^6)=0.5$. Leaf nodes ($s^4, s^5, s^6, s^7$) are terminal states where the agent will receive a reward: $R(s^2,a^1,s^4)=+10$, $R(s^2,a^2,s^5)=-10$, $R(s^2,a^2,s^6)=+20$, and $R(s^3,a^2,s^7)=-20$. Color shading indicates the decisions that different agents will make after a single visit to every possible trajectory (orange for episodic control and grey for 1-step Q-learning). We can see that after the agent discovers all possible trajectories, the episodic control agent is already able to follow a sub-optimal trajectory starting from the initial state. However, the 1-step Q-learning agent only back-propagates the reward signal from terminal states to their predecessors, which means the 1-step Q-learning agent still is not able to make decisions in the initial state. The optimal policy for this MDP is to take $a^1$ in $s^1$ and take $a^1$ in $s^2$ as well. By definition, episodic control will converge to a sub-optimal policy that always takes $a^2$ in $s^2$ (shown in Eq.~\ref{eq:q_ec}). With more updates, 1-step Q-learning will learn the optimal state-action value (shown in Eq.~\ref{eq:q_rl}) for each state-action pair, which will result in the optimal policy. 

\begin{equation}
    Q^{ec}(s^2,a^1)=10, Q^{ec}(s^2,a^2)=20
    \label{eq:q_ec}
\end{equation}
\begin{equation}
    Q^*(s^2,a^1)=10, Q^*(s^2,a^2)=5
    \label{eq:q_rl}
\end{equation}

Thus, we conclude that episodic control is fast but sub-optimal, and reinforcement learning (1-step Q-learning in our case) is slow but optimal. There should exist an intersection point between these two different methods where reinforcement learning surpasses episodic memory.
%, if we can combine these two methods and somehow detect such an intersection point, we will benefit from both sides. 
One might ask why we do not compare with (full-episode) Monte-Carlo backup. The 1-step back-up can utilize off-policy learning to learn from data collected by other policies, which is more data efficient. We therefore aim to combine the fast learning of EC with the data efficiency of 1-step off-policy learning. 

\subsection{Switching}
\label{subsec:switch}
The previous example highlighted that EC may learn a solution faster, while RL may eventually converge to a better solution. Therefore, we ideally want a switching mechanism, that transitions from EC action selection to RL action selection. We need to decide both {\bf when} and {\bf how} to switch between the two different `memories'. 

Regarding the `when', we propose to decide which memory to use for every episode. This way, we ensure that action selection within the episode stays consistent. To determine which memory we will use in a particular episode, we need to define a probability $p^{ec}$ that specifies the probability that we will use EC in the next episode. Obviously, we then select {\it 2M-RL} with probability $1-p^{ec}$. Since we favor the use of {\it 2M-EC} at the beginning and {\it 2M-RL} near the end, we want to gradually decay $p^{ec}$ from a high value to a lower value according to the equation:

\begin{equation}
    p^{ec} \gets p^e + (p^s-p^e) \cdot e^{-i/\tau}
    \label{eq:p}
\end{equation}

where $p^s$ and $p^e$ are starting value and end value of $p^{ec}$, $i$ is the number of steps the agent takes so far and $\tau$ is the temperature that controls the speed of decay. Smaller $\tau$ decays $p^{ec}$ faster while lager $\tau$ decays $p^{ec}$ slower. In the ablation experiments, we also experiment with different scheduling mechanisms. 

During evaluation, we exploit knowledge stored in both memories in a greedy manner. However, we still need to decide which memory to use during evaluation. The scores {\it 2M-RL} and {\it 2M-EC} obtained during training are used as metrics to evaluate their respective performances, and the memory with the higher recent score is selected for action selection during evaluation. More specifically, we keep track off cumulative rewards of both memories during training, and the score ($s^{rl}$ and $s^{ec}$ for {\it 2M-RL} and {\it 2M-EC}, respectively) is defined as the average cumulative reward over the last $n$ episodes per memory method. We fix $n=50$ in this work. Thus, the {\it 2M} agent will choose among two memories using Eq.~\ref{eq:switch_eval}.
\begin{flalign}
 \textit{2M} \gets
    \begin{cases}
      \textit{2M-RL} & \text{if $s^{rl} \geq s^{ec}$}, \\
      \textit{2M-EC} & \text{otherwise}
    \end{cases} \nonumber \\
%\frac{1}{n} \sum_{i=1}^n R(\tau_i^{EC})
    \label{eq:switch_eval}
\end{flalign}
\subsection{Learning}
\label{subsec:learning}
To foster mutual improvement of the two memories, the collected data is shared between {\it 2M-EC} and {\it 2M-RL}. Regardless of which memory the data is collected by, it is used to update {\it 2M-EC} according to Eq.~\ref{eq:ec_update}, ensuring that {\it 2M-EC} is always up to date. On the other hand, since we use off-policy reinforcement learning methods, data collected by {\it 2M-EC} can also be used to update {\it 2M-RL}. This is implemented by maintaining a replay buffer, and all data collected during the training will be pushed into it. {\it 2M-RL} is then trained (Eq.~\ref{eq:dqn}) every $x$ timesteps (we use $x=10$) by sampling minibatches from the buffer. It should be noted that the value of $p^{ec}$ indirectly determines the amount of data in the replay buffer originating from {\it 2M-EC}, and thereby the proportion of such data used for training {\it 2M-RL}. 

%%%%%%%%%%%%%%%%%%%%%%%%%%%%%%%%%%%%%%%%%%%%
\section{Experiments}
We first present experimental results on a simple WindyGrid environment to demonstrate the efficiency of the proposed {\it 2M} agent. Subsequently, we perform extensive experiments on five MinAtar~\cite{young19minatar} games, namely Breakout, SpaceInvaders, Asterix, Seaquest, Freeway, as illustrated in Fig.~\ref{fig:minatar_env}. These games present diverse challenges, such as exploration (Seaquest and Freeway), classical control tasks (Breakout and SpaceInvaders), and so on. MinAtar is a simplified version of Atari Learning Environment~\cite{bellemare2013arcade}, which simplifies representations of games while still capturing general mechanics. This allows the agent to focus on behavioural challenges rather than representation learning. Finally, an ablation study is conducted to investigate the crucial choices of the proposed method.

\subsection{Proof of Concept Under Tabular RL Setting}
\begin{figure}[!htb]
    \centering
    \includegraphics[scale=0.3]{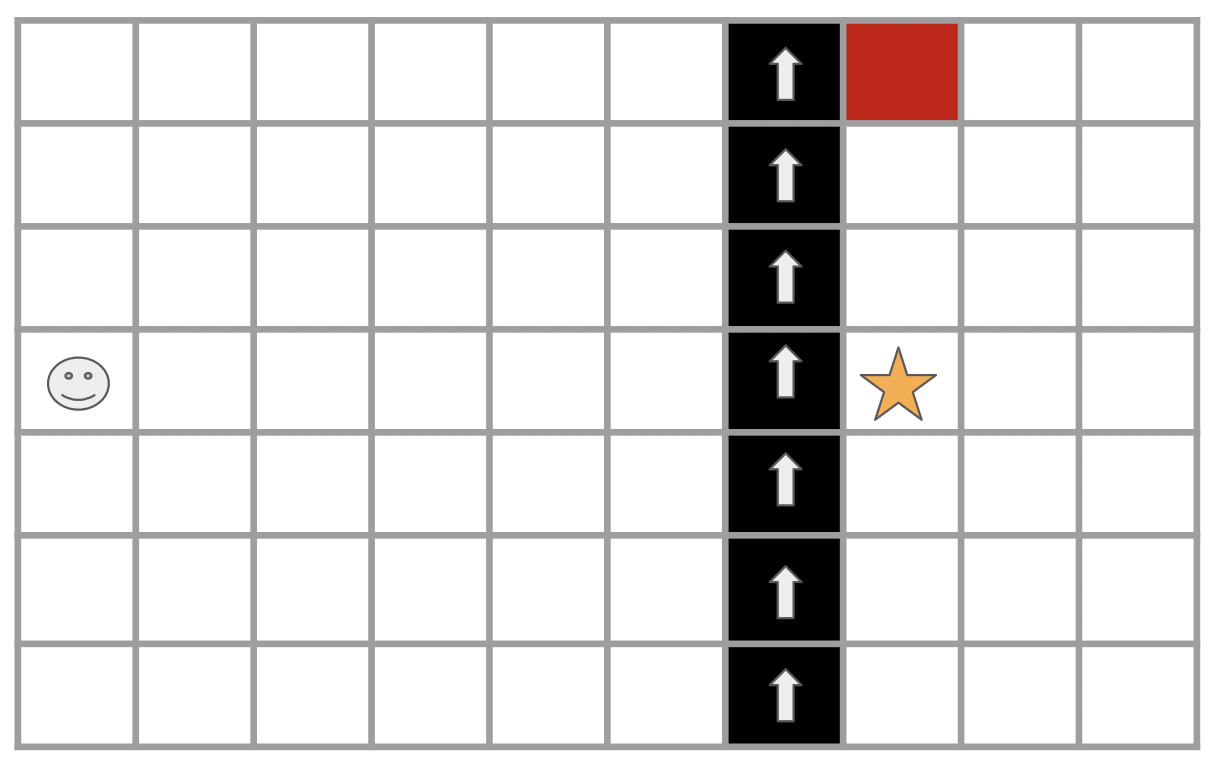}
    \caption{An illustration of the WindyGrid environment. The agent (smile face) needs to navigate from the starting point to the given goal (star). There is the stochastic wind that will (black column with up arrows) blow the agent up and the agent gets a penalty when it reaches the trap (red cell).}
    \label{fig:windy}
\end{figure}

The {\it 2M} agent integrates tabular 1-step Q-learning with episodic control to solve a WindyGrid task (shown in Fig.~\ref{fig:windy}). The WindyGrid instance we use here is $7 \times 10$ large, with a stochastic wind in the $7^{th}$ row and a trap where the agent will get a large penalty. The agent needs to navigate from the initial state $(3, 0)$ to the terminal state $(3, 7)$.
\begin{figure}[!htb]
    \centering
    \includegraphics[scale=0.17]{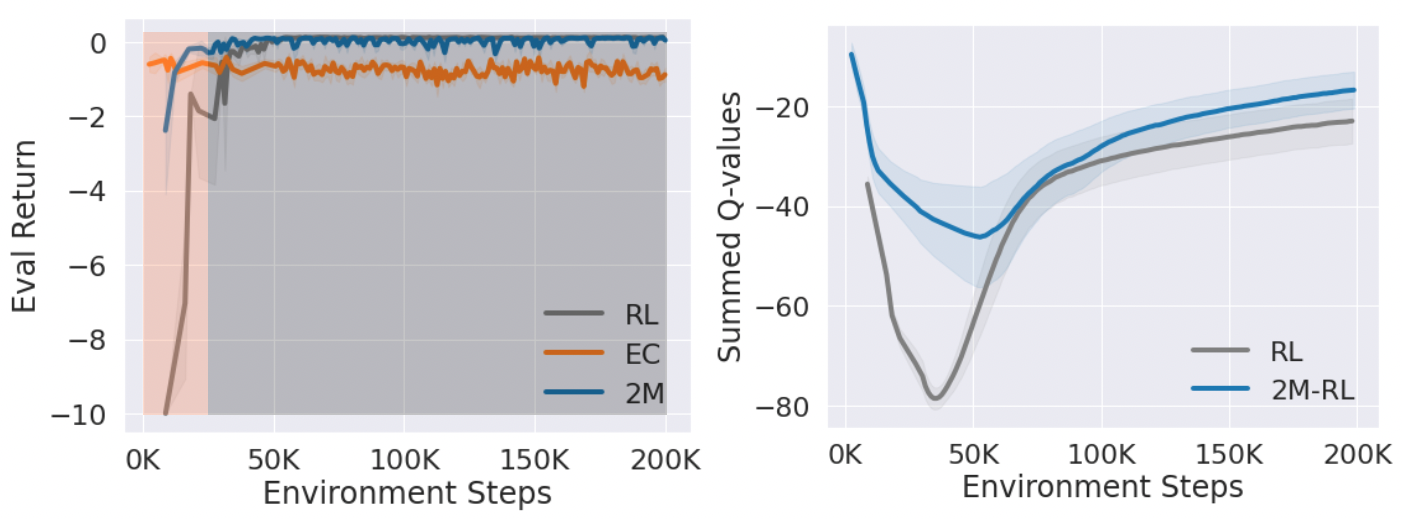}
    \caption{Results on WindyGrid under tabular settings. \textbf{Left}: Evaluation returns of different agents, EC learns very fast but converges to local optima while RL learns slowly but converges to global optima. 2M learns faster (compare to RL) at the beginning and has better asymptotic performance (compare to EC). Colors in the background indicate memories used during evaluation, orange and grey represent the use of {\it 2M-EC} and {\it 2M-RL}. \textbf{Right}: Learned Q-values summed across the entire state-action space, {\it 2M-RL} learns Q-values faster than pure RL which is trained using monotonic RL data.}
    \label{fig:tab}
\end{figure}
The results presented in Fig.~\ref{fig:tab} demonstrate the performance of various agents. The left figure shows returns the agent obtains during the evaluation, while the right figure shows learned Q-values summed across the whole state-action space. The orange line represents the {\it EC} agent, which achieves quick learning and decent performance from the start but only has sub-optimal asymptotic performance. In contrast, the grey line corresponds to the {\it RL} agent, which initially performs poorly but gradually improves and eventually converges to the optimal solution. The blue line corresponds to the {\it 2M} agent, which learns faster at the beginning compared to {\it RL} and achieves better asymptotic performance compared to {\it EC}. The background colors indicate the use of different memories by the {\it 2M} agent during the evaluation, with the agent using {\it 2M-EC} (orange) for evaluation at the beginning and then switching to {\it 2M-RL} (grey) after approximately $25k$ steps.  

Since the memories {\it 2M-RL} and {\it 2M-EC} share data, and {\it 2M-RL} is trained on data collected by both of them, we investigate whether this approach has a positive impact compared to training solely on data collected by pure {\it RL}. 
As Varun et al.~\cite{kompella2022event} demonstrated theoretically and experimentally, learning from data correlated to the optimal policy will results in lower complexity (which is an intuitive result). If we assume the EC data is at least somewhat correlated to the optimal policy, we expect that use of EC data in RL updates will lead to faster learning. We experimentally test this idea in Fig.~\ref{fig:tab}. The right panel of Fig.~\ref{fig:tab} shows that using mixed data to train the {\it RL} agent ({\it 2M-RL}) can indeed result in faster learning of state-action values compared to using data solely collected by {\it RL}. This suggest EC data may actually improve RL sample efficiency.  

\subsection{Results on MinAtar Games}
Next, we perform experiments on five MinAtar games, which also vary in the amount of stochasticity. We optimize hyper-parameters over the values shown in Tab.~\ref{tab:hyperpara}, where the settings used for the final results are highlighted in bold.

\begin{table}[!htb]
    \caption{Tuned hyper-parameters, where the ones we use for the final experiments are highlighted in bold. $p^s$ is the starting value from Eq.\ref{eq:p}}
    \centering
    \begin{tabular}{|c|c|}
        \hline
         $\epsilon$ (for exploration)  & ${\bf 0.1}, 0.9$ \\
         \hline
         $k$ (for $k$-NN in EC) & $1, {\bf 3}, 10$ \\
         \hline
         $p^{s} \to p^{e} $ (for switching) & $\bf 0.9 \to 0.1, 0.1 \to 0.1, 0.1 \to 0.9$ \\
         \hline
         learning rate & $0.001, \textbf{0.0001}$ \\
         \hline
    \end{tabular}
    \label{tab:hyperpara}
\end{table}
\begin{figure*}[!htb]
    \centering
    \includegraphics[scale=0.4]{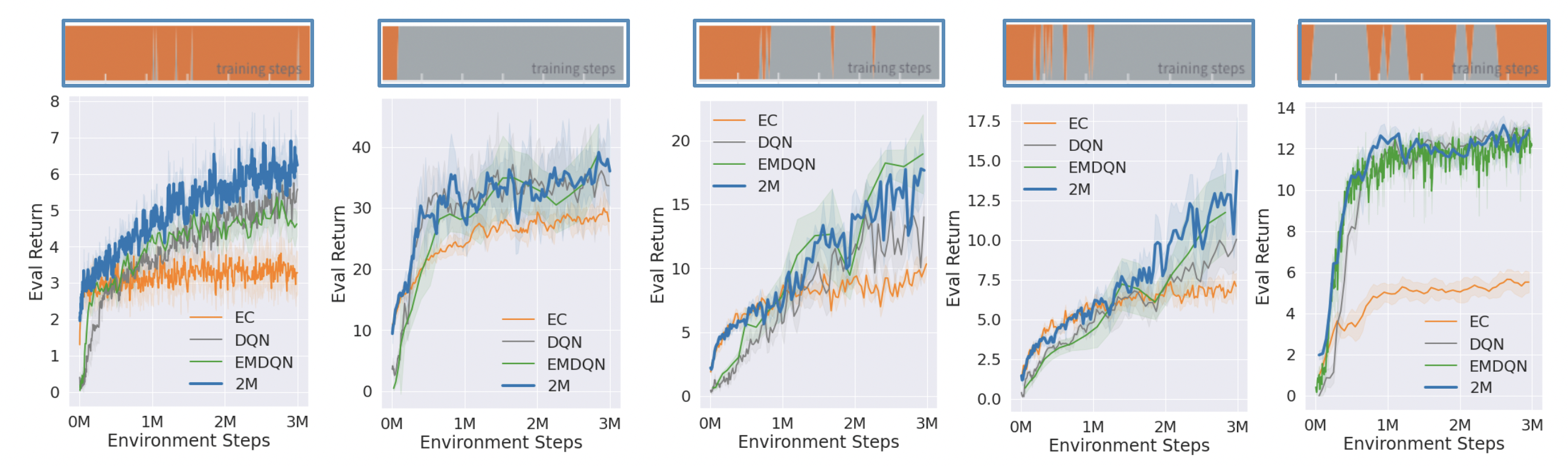}
    \caption{Results on MinAtar games: Breakout, SpaceInvaders, Asterix, Seaquest, Freeway. The top row shows the relevant memory that the {\it 2M} agent chooses for evaluation: orange represents {\it 2M-EC} and grey represents {\it 2M-RL}. The bottom row shows the returns obtained during training (running independent evaluation epsisodes). The {\it 2M} agent either outperforms or is on par with all baseline comparisons (EC, DQN and EMDQN). EC generally learns fast, but then reaches a plateau. {\it 2M} learns equally fast initially, but then adopts the better long-term performance of RL. }
    \label{fig:MinAtarResults}
\end{figure*}

In the results presented in Fig.~\ref{fig:MinAtarResults}, the top rows depict the various memories the {\it 2M} agent selected during the evaluation, whereas the bottom rows display the returns agents obtain. The performance of {\it EC} (orange lines) and {\it DQN} (grey lines) is consistent with the findings observed in the toy example, where {\it EC} learns quickly at the beginning but converges to sub-optimal solutions while {\it DQN} learns slowly but has better asymptotic performance. {\it 2M} agents (blue lines) perform the best (comparably to {\it EC}) at the beginning on all five games, and in most games, they also exhibit better asymptotic performance (or at least equally good performance) compared to other baselines (except maybe for Asterix). In Asterix, the {\it 2M} agent underperforms {\it EMDQN} (green lines), but is still better than {\it DQN} and {\it EC}. 

During the evaluation, most {\it 2M} agents demonstrate a preference for using {\it 2M-EC} initially and then switching to {\it 2M-RL} over time. In Freeway, the agent continually switches between {\it 2M-EC} and {\it 2M-RL}, suggesting a mutually beneficial relationship between these two memories. However, in Breakout, the agent consistently favors to use {\it 2M-EC}. This may be attributed to the fact that stochastic dynamic transitions have a less pronounced impact on performance in this game, with the most critical time step being the one at which the paddle must bounce the ball. We hypothesize {\it 2M-RL} can help {\it 2M-EC} escape local optima and then {\it 2M-EC} can rapidly learn improved solutions and continue to progress. To test this hypothesis, we need to check whether a {\it 2M} agent that does not share collected data between both memories indeed 1) has worse performance, and 2) prefers RL in the long run, instead of getting to a more optimal EC solution.

\begin{figure}[!htb]
    \centering
    \includegraphics[scale=0.31]{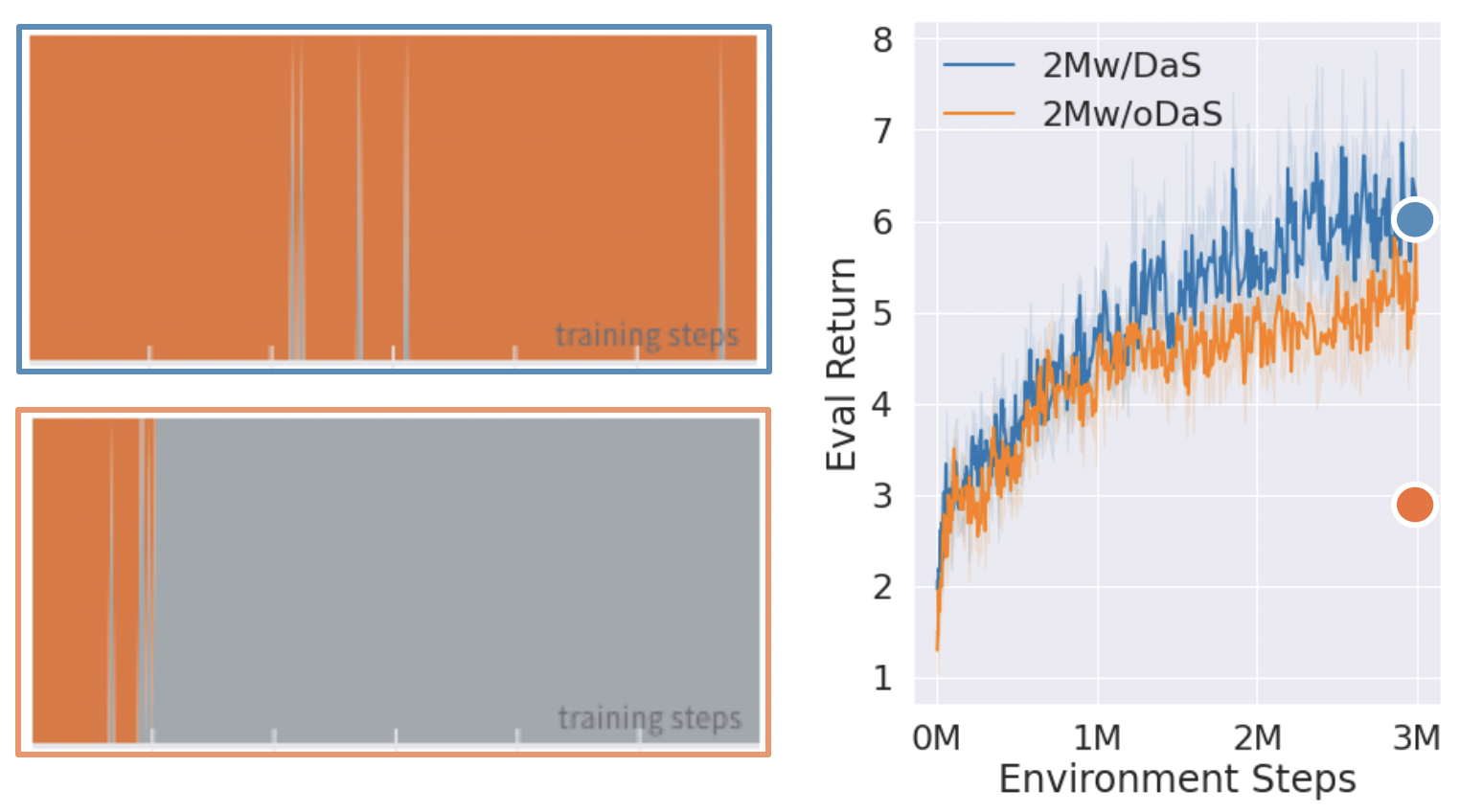}
    \caption{{\bf Left}: Switching schedule of {\it 2M} agent with data sharing (top) and without data sharing (bottom) during evaluation. {\bf Right}: Returns two different {\it 2M} agents are able to get in evaluation. The final performance of {\it 2M-EC} is represented by coloured circles. We see that with data sharing, EC reaches a return of 6, while without data sharing, EC only manages to reach a return of 3.}
    \label{fig:breakout}
\end{figure}

 Fig.~\ref{fig:breakout} shows the performance of an {\it 2M} agent with data sharing enabled and disabled. With data sharing (2Mw/DaS), the agent mostly prefers to use EC during evaluation (top-left figure), as expected from Fig. \ref{fig:MinAtarResults}. When we deactivate data sharing (2Mw/oDaS, two memories are only trained using data collected by the corresponding memory separately), the {\it 2M} agent only prefers {\it 2M-EC} at the beginning and then sticks to {\it 2M-RL} (bottom-left graph of the figure). The performance graph on the right of the figure confirms these results. Without data sharing, 2M does not reach the same performance (blue line stays above the orange line). The circles show the performance of {\it 2M-EC} at the end of training for both methods. Without data sharing, {\it 2M-EC} (the orange circle in Fig.~\ref{fig:breakout}) converges to a sub-optimal solution. With data sharing enabled, {\it 2M-EC} (the blue circle in Fig.~\ref{fig:breakout}) has a way higher performance. This observation provides evidence to support the aforementioned notion that {\it 2M-RL} and {\it 2M-EC} complement eachother. 
 
\subsection{Ablation Study} 
In this section, we conduct two groups of ablation experiments to study the design choices in our work. First, we would like to investigate the impacts of data sharing. Deactivating \textit{data sharing} (2Mw/oDS), resulting in {\it 2M-RL} being solely trained on data collected by {\it 2M-RL} and {\it 2M-EC} being solely trained on data collected by {\it 2M-EC}. This transforms our proposed method becomes a `scheduler' that schedules the training between two distinct models and uses the better one for evaluation. Second, we aim to study different ways of scheduling $p^{ec}$. Specifically, we examine three different scheduling approaches: decayed scheduling (2Mw/DS), constant scheduling (2Mw/CS) and increased scheduling (2Mw/IS).

\begin{figure}[!htb]
    \centering
    \includegraphics[scale=0.34]{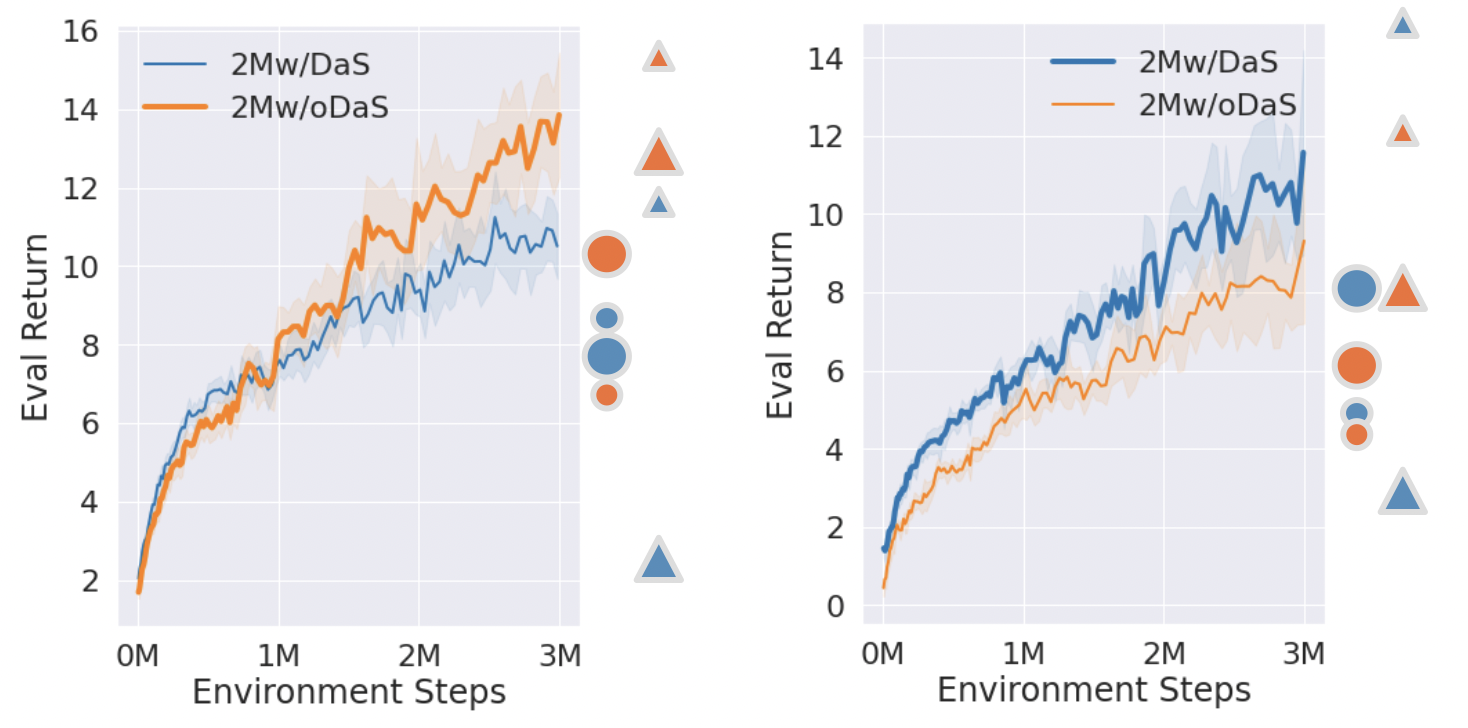}
    \caption{Performance of 2M agents with and without data sharing on Asterix (left) and Seaquest (right). Results are averaged over 2 different settings, one is with larger $p^{ec}$ and one is with smaller $p^{ec}$. Data sharing has different impacts on these two games. Circles represent the final performance of {\it 2M-EC} while triangles represent the performance of {\it 2M-RL}. The larger size means the larger value of $p^{ec}$ during the training.}
    \label{fig:as_ds}
\end{figure}

Intuitively, data sharing can be helpful since each memory will get different data from another memory, hopefully, they could also learn from each other. It should result in a better performance compared to only training the agent on its own collected data. In fact, such sharing has different impacts on different games, shown in Fig.~\ref{fig:as_ds}. In Asterix, data sharing improves {\it 2M} agent's performance, and harms the performance of the agent in Seaquest. To understand the reasons for these opposite impacts, we separately track the performance of {\it 2M-EC} and {\it 2M-RL} during the training, and final performance are represented by circles and triangles in Fig.~\ref{fig:as_ds} and larger size represents more use of {\it 2M-EC} (larger $p^s$) during the training. In Asterix, data sharing pulls down the performance of {\it 2M-RL} (blue triangles are always below orange ones), and training {\it 2M-RL} on more data collected by {\it 2M-EC} leads to even worse performance (the large blue triangle is way below the large orange one), indicating that data collected by {\it 2M-EC} is actually harmful for training {\it 2M-RL} in this game. Conversely, in Seaquest, data sharing improves the performance of {\it 2M-EC} (blue circles are always above orange ones), which again indicates that {\it 2M-RL} can help {\it 2M-EC} escape from local optima. However, overusing data collected by {\it 2M-EC} to train {\it 2M-RL} also leads to worse performance (the large blue triangle is way below the large orange one). All in all, {\it 2M-RL} should not use too much data collected by {\it 2M-EC}. Although we show that such training is helpful to learn the optimal state-action values when collected data is correlated to the optimal policy, this assumption is not always satisfied. Meanwhile, {\it 2M-RL} can help {\it 2M-EC} escape from local optima but not always. We presume it helps when stochastic transitions have fewer negative impacts, then {\it 2M-EC} is able to catch improved solutions provided by {\it 2M-RL}. For example, in Asterix, there are many enemies and the agent dies immediately when it touches enemies, meaning the agent will more likely die if a single wrong (sub-optimal) action is made. Therefore, once {\it 2M-EC} discovers a sub-optimal trajectory with a very high return (luckily manage to survive for a long time, like ending up with $s_6$ in the motivating example in Fig.~\ref{fig:motivation}) with a tiny probability, it will stick to it. Then it is difficult for {\it 2M-EC} to escape from local optima even though improved solutions are provided. We leave more systematic investigations on this phenomenon for future work.

\begin{figure}[!htb]
    \centering
    \includegraphics[scale=0.55]{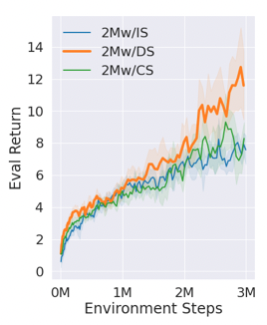}
    \caption{Performance of {\it 2M} agents with different scheduling mechanisms on Seaquest. The one with the decayed scheduling mechanism works the best while other two mechanisms have similar performance.}
    \label{fig:as_ec}
\end{figure}

Next we examine how different scheduling mechanisms affect performance. The {\it 2M} agent with decayed scheduling ({\it 2Mw/DS}) will initially give a higher preference to use {\it 2M-EC} for data collection during the training, then gradually shifts towards {\it 2M-RL}. On the contrary, increased scheduling ({\it 2Mw/IS}) will start with a strong preference for using {\it 2M-RL}, then gradually switch to {\it 2M-EC}. Constant scheduling maintains a constant preference for {\it 2M-EC} and {\it 2M-RL} throughout the training process. Given that {\it 2M} agents with {\it 2Mw/DS} perform the best or one of the best on all games, we only present the performance on Seaquest as a representative in Fig.~\ref{fig:as_ec}. The results demonstrate that the agent with {\it 2Mw/DS} outperforms the other two scheduling mechanisms explicitly. 

Our experimental results demonstrate that the {\it 2M} agent surpasses its constituent components, i.e. a pure reinforcement learning approach (Q-learning in tabular settings and DQN~\cite{mnih2013atari} in deep RL settings) and a pure episodic memory approach (MFEC~\cite{blundell2016model} in both tabular and deep RL settings). Furthermore, our proposed method exhibits better performance than a state-of-the-art memory-augmented reinforcement learning method EMDQN~\cite{lin2018episodic}. Since the proposed framework allows for the integration of various pure EC and RL methods, we only compare the performance of the {\it 2M} agent with those methods that are integrated into it in this work. Therefore, we do not compare our results with other pure episodic memory and reinforcement learning approaches. Lastly, we conduct ablation studies to investigate the impact of two essential design choices: the utilization of data sharing for the mutual improvement of the memories, and the scheduling of $p^{ec}$ for switching between the two memories.
%%%%%%%%%%%%%%%%%%%%%%%%%%%%%%%%%%%%%%%%%%%%
\section{Conclusion and Future Work}
In this work, we proposed a novel approach, termed {\bf Two-Memory ({\bf \it 2M}) reinforcement learning} agent, which integrates two distinct learning methods, namely non-parametric episodic memory and (parametric) reinforcement learning. This approach capitalizes on the advantages of both methods by leveraging the episodic memory's rapid starting and the reinforcement learning's superior generalization and optimality to create an agent that achieves higher data efficiency than the individual methods. Experimental results show that {\it 2M} matches or outperforms both DQN and EMDQN on a range of MinAtar games. In addition, we show that these two distinct memory modules may complement eachother, leading to even better final performance. 

For future work, it would be interesting to automatically determine when data sharing is useful, for example based on the discrepancy between both memories. Another clear direction to improve the {\it 2M} agent could be an adaptive scheduling mechanism to switch between {\it 2M-EC} and {\it 2M-RL}, instead of the hand-designed decay schedules used in this work. Moreover, combining stronger episodic memory methods (such as NEC) with off-policy reinforcement learning methods could lead to further improvements in performance. In our work, we uniformly replay data from the replay buffer to train the {\it 2M-RL} component, but a more sophisticated replay strategy, such as prioritized experience replay (PER), may further enhance performance. Overall, our proposed approach provides a general framework for combining two fundamentally different approaches to sequential decision-making, combining their respective strengths.
\bibliographystyle{IEEEtran}
\bibliography{IEEEabrv,ref}

\end{document}